\documentclass[journal]{IEEEtran}

\ifCLASSINFOpdf
\else
   \usepackage[dvips]{graphicx}
\fi
\usepackage{url}

\usepackage{graphicx}
\usepackage{amsmath}
\usepackage{amsfonts}
\usepackage{amssymb}
\usepackage{float}
\begin{document}

\title{Supervised Image Translation from Visible to Infrared Domain for Object Detection}

\author{Prahlad Anand, Qiranul Saadiyean, Aniruddh Sikdar, Nalini N and Suresh Sundaram
\thanks{Prahlad Anand is with the School of Computer Science and Engineering, Vellore Institute of Technology, Vellore. (e-mail: prahlad.anand2020@vitstudent.ac.in).}
\thanks{Qiranul Saadiyean is with the Department of Aerospace Engineering, Indian Institute of Science, Bangalore. (e-mail: qiranuls@iisc.ac.in).}
\thanks{Aniruddh Sikdar is with the Robert Bosch Centre for Cyber-Physical Systems, Indian Institute of Science, Bangalore. (e-mail: anirudhss@iisc.ac.in}
\thanks{Nalini N is with the School of Computer Science and Engineering, Vellore Institute of Technology, Vellore. (e-mail: nalini@vit.ac.in).}
\thanks{Suresh Sundaram is with the Department of Aerospace Engineering, Indian Institute of Science, Bangalore. (e-mail: qiranuls@iisc.ac.in).}}

\maketitle

\markboth{}
{Shell \MakeLowercase{\textit{et al.}}: Bare Demo of IEEEtran.cls for IEEE Journals}

\begin{abstract}
This study aims to learn a translation from visible to infrared imagery, bridging the domain gap between the two modalities so as to improve accuracy on downstream tasks including object detection. Previous approaches attempt to perform bi-domain feature fusion through iterative optimization or end-to-end deep convolutional networks. However, we pose the problem as similar to that of image translation, adopting a two-stage training strategy with a Generative Adversarial Network and an object detection model. The translation model learns a conversion that preserves the structural detail of visible images while preserving the texture and other characteristics of infrared images. Images so generated are used to train standard object detection frameworks including Yolov5, Mask and Faster RCNN. We also investigate the usefulness of integrating a super-resolution step into our pipeline to further improve model accuracy, and achieve an improvement of as high as 5.3\% mAP.
\end{abstract}


\IEEEpeerreviewmaketitle

\section{Introduction}

\IEEEPARstart{I}{nfrared} imaging has increased in popularity as a tool in computer vision tasks, offering unique advantages over conventional visible light imaging, particularly in scenarios where visibility is limited or compromised \cite{c1,c2}. The use of infrared images has significantly enhanced various computer vision tasks, including object detection, due to its ability to capture thermal signatures and penetrate adverse environmental conditions such as fog, smoke, and darkness. However, the scarcity of annotated datasets and the domain gap between visible and infrared imagery poses significant challenges to computer vision models.

Image fusion is popularly used to combine the spectral information from both infrared and visible modalities to create a single fused image that enhances the overall visibility and information content of the scene \cite{c5,c6,c7}. However, this approach can lead to information loss, particularly in regions where one modality provides more discriminative features than the other. Further, the lack of quantitative metrics for evaluating fusion quality further complicates the selection of an optimal fusion method.
The performance of fusion techniques heavily depends on the diversity and representativeness of the training dataset. Most importantly, image fusion techniques rely on the assumption that both visible and infrared image data is available for training \cite{c8,c9}, which is not necessarily true.

Generative adversarial network architectures have become popular for translating visible images to infrared images \cite{c10}. In particular, Ozkanöglu et al. \cite{c11} attempt to focus on the structural similarity (SSIM) metric during the image translation process. While SSIM is commonly used as a perceptual similarity metric for evaluating image quality, over-reliance within the InfraGAN architecture leads to lower performance and artifact generation when applied to datasets with complex structures and diverse visual characteristics.\\
Isola et al. introduced Pix2Pix \cite{c12}, leveraging conditional adversarial networks (GANs) to learn mapping functions between image pairs. However, Pix2Pix tends to produce results with limited texture details and suffers from mode collapse issues. Liu et al. \cite{c13} addresses the limitations of Pix2Pix by introducing a few-shot learning paradigm, where the model can adapt to new domains with minimal training data by decoupling content and style representations. Building upon FUNIT, COCO-FUNIT, proposed by Lee et al. \cite{c14}, incorporates contrastive learning techniques to enhance the disentanglement of content and style representations. By explicitly aligning the latent spaces of different domains, COCO-FUNIT achieves superior translation performance through better image structure and texture preservation.
Li et al. introduce a novel attention mechanism \cite{c10} that focuses on both semantic and instance-level information during image translation. Zhou et al. \cite{c15} introduce a branch-structured network architecture that facilitates bidirectional translation between visible and infrared domains, incorporating cross-domain consistency constraints and cycle consistency loss. Chen et al. \cite{c16} leverage intermediate sketch representations to bridge the semantic gap between the visible and infrared domains.\\
Pix2PixHD \cite{c17} extends the Pix2Pix framework by introducing a hierarchical multi-scale generator architecture, enabling high-resolution image translation. By incorporating global and local adversarial losses, Pix2PixHD produces visually appealing infrared images with fine-grained details and sharp textures, making it suitable for various applications requiring high-fidelity image translation.
Early image fusion methods utilised deep neural networks for feature extraction or to learn weights \cite{c4,c18}. Liu et al. \cite{c19} cascaded two pretrained CNNs, one for feature and the other for weights learning. End-to-end architectures can generate a plausible fused image by one set of network parameters. Li et al. \cite{c20} introduced a residual fusion network to learn enhanced features in a common space, yielding structurally consistent results amenable to human inspection.
Ma et al. introduced adversarial training between the fused and visible in order to enhance texture details \cite{c21}. However, it is likely that a significant amount of information from infrared images is lost due to the signal adversarial mechanism. Ma et al. \cite{c22} apply an identical adversarial strategy to both the visible and infrared, which partially compensates the infrared information. Unfortunately, virtually all approaches fail to relate the differing characteristics of visible and infrared images.

\begin{figure*}
    \centering
    \includegraphics[scale=0.45]{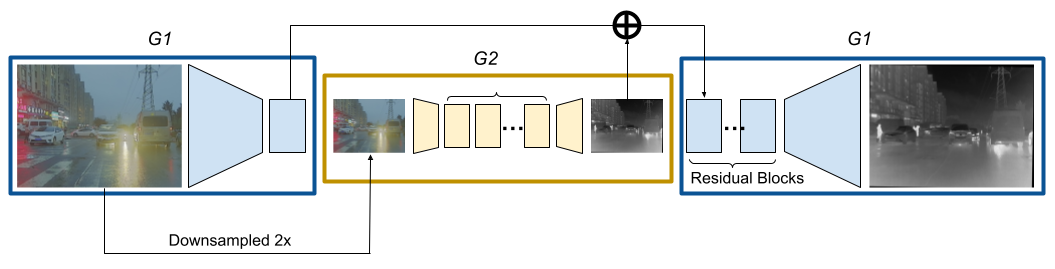}
    \caption{Network architecture of the generator. A residual network $G_{1}$ is first trained on lower resolution images. A second residual network $G_{2}$ is then appended to $G_{1}$ and the two networks are trained jointly on high-resolution images. Specifically, the input to the residual blocks in $G_{2}$ is the element-wise sum of the final feature map from $G_{1}$ and the feature map from $G_{2}$}
    \label{fig:Arch}
\end{figure*}

\section{Methodology}

We utilise a conditional adversarial framework for generating high-resolution and realistic generated infrared images from co-registered source visible and infrared images. First, an overview of the baseline pix2pix model is provided. Subsequently, we cover the improved objective function and network design used to improve results, particularly for high-resolution images. Additionally, to further improve the image quality, an instance-level object semantic information and feature embedding scheme is used to better handle the multi-modal image synthesis.

\subsection{Pix2pix Baseline}
The pix2pix method \cite{c12} is a typical conditional GAN framework for image translation, consisting of a generator $G$ and a discriminator $D$. For our task, the objective of the generator $G$ is to translate visible images to corresponding infrared images, while the discriminator $D$ aims to distinguish real infrared images from the translated ones. The framework operates in a supervised setting. The training dataset can thus be described as consisting of paired images $\left\{\left(\mathbf{s}_{\mathbf{i}}, \mathbf{x}_{\mathbf{i}}\right)\right\}$, where $\mathbf{s}_{\mathbf{i}}$ is a visible image and $\mathbf{x}_{\mathbf{i}}$ is its co-registered infrared image. All Conditional GANs model a minimax game to learn model the conditional distribution of infrared images given the input visible images:

\begin{equation}
\min _{G} \max _{D} \mathcal{L}_{\mathrm{GAN}}(G, D)
\\
\end{equation}

where the objective function $\mathcal{L}_{G A N}(G, D)$ is given by

\begin{equation}
\mathbb{E}_{(\mathbf{s}, \mathbf{x})}[\log D(\mathbf{s}, \mathbf{x})]+\mathbb{E}_{\mathbf{s}}[\log (1-D(\mathbf{s}, G(\mathbf{s}))]
\\
\end{equation}

With U-Net \cite{c23} as the generator, the patch-based fully convolutional discriminator receives as input a channelwise concatenation of the visible image and the co-registered infrared image. As the resolution of the generated images on M3FD \cite{c5} is up to $256 \times 256$ and the  generated highresolution images of insufficient quality for downstream tasks (object detection), we adapt an improved model for our task.

\subsection{Modifications}
The pix2pix framework is improved by using a coarse-to-fine generator, a multi-scale discriminator architecture, and a robust adversarial learning objective function, as per Wang et al. \cite{c17}.

\subsubsection{Coarse-to-fine Generator}
The generator is composed of two sub-networks: $G_{1}$ and $G_{2}$. $G_{1}$ is termed as the global generator network and $G_{2}$ as the local enhancer network. The tuple $G=$ $\left\{G_{1}, G_{2}\right\}$ then describes the generator as visualized in Fig. \ref{fig:Arch} The global generator network operates at a resolution of $1024 \times 512$, and the local enhancer network outputs images with resolution $4 \times$ that of the previous one (twice along each image dimension). Additional local enhancer networks could be made use of to generate image of higher resolution. For example, the output image resolution of the generator $G=\left\{G_{1}, G_{2}\right\}$ is $2048 \times 1024$, and the output image resolution of $G=\left\{G_{1}, G_{2}, G_{3}\right\}$ is $4096 \times 2048$.

The global generator is built on Johnson et al.'s neural style transfer architecture \cite{c24}, successful on images of size up to $512 \times 512$. It comprises 3 principal sub-networks: a convolutional downsampling network $G_{1}^{(F)}$, residual blocks $G_{1}^{(R)}$ \cite{c25}, and a transposed convolutional upsampling network $G_{1}^{(B)}$. Visible images of resolution $1024 \times 512$ are passed through the 3 components sequentially to output infrared images of equal resolution.

The local enhancer network itself similarly comprises 3 sub-networks: convolutional front-end $G_{2}^{(F)}$, residual blocks $G_{2}^{(R)}$, and transposed convolutional back-end $G_{2}^{(B)}$. The resolution of the visible image to $G_{2}$ is twice that of the global generator along both diensions ($2048 \times 1024$). However, in contrast to the global generator network, the input to the residual block $G_{2}^{(R)}$ is the element-wise sum of both the output feature map of $G_{2}^{(F)}$, as well as the final feature map of the upsampling network of the global generator network $G_{1}^{(B)}$. This aids in integrating the global information from $G_{1}$ to $G_{2}$ \cite{c17}.

During training, the global generator and local enhancer are trained sequentially, in the order of their resolutions. All networks are then fine-tuned together. This generator design effectively aggregates global and local information for the image generation and translation. Similar multi-resolution pipelines are well established in computer vision \cite{c26} with two-scale architectures often proving sufficient \cite{c27}.

\subsubsection{Multi-scale discriminator} High-resolution image synthesis is difficult to resolve with respect to the design of the GAN discriminator. A larger receptive field for the discriminator would aid in differentiating high-resolution real and generated images. This might be achieved through either of a deeper network or larger convolutional kernels, both of which would increase the network capacity (and consequently require greater memory for training) and potentially cause overfitting.

Thus, 3 separate discriminators $D_{1}, D_{2}$ and $D_{3}$ that have  identical network structure but operate at different image scales are used. Real and generated high-resolution images are downsampled by a factor of 2 and 4 to create an image pyramid of 3 scales in total. Then individual discriminators $D_{1}, D_{2}$ and $D_{3}$ are  trained to differentiate real and generated images at corresponding scales. The discriminator that operates at the coarsest scale has the largest receptive field, thus having a more global view of the image, potentially guiding the generator to generate globally consistent images. Similarly, but at the finest scale, the relevant discriminator directs the generator to produce finer details. Extending a low-resolution model to a higher resolution only requires adding a discriminator at the finest level, rather than retraining from scratch, making training the coarse-to-fine generator easier. If the generator is trained without the multi-scale discriminators, it is observed that many repeated patterns often appear in the generated images \cite{c17}.

With the discriminators, the learning problem in Eq. 1 then becomes a multi-task learning problem of

\begin{equation}
\\\min _{G} \max _{D_{1}, D_{2}, D_{3}} \sum_{k=1,2,3} \mathcal{L}_{\mathrm{GAN}}\left(G, D_{k}\right)
\\
\end{equation}

\subsubsection{Improved adversarial loss} The GAN loss in Eq. 2 is improved by incorporating a feature matching loss based on the discriminator. This loss stabilizes the training as the generator has to produce natural statistics at multiple scales. Specifically, features are extracted from multiple layers of the discriminator and learn to match these intermediate representations from the real and the generated image. The $i$ th-layer feature extractor of discriminator $D_{k}$ is denoted as $D_{k}^{(i)}$ (from input to the $i$ th layer of $D_{k}$ ). The feature matching loss $\mathcal{L}_{\mathrm{FM}}\left(G, D_{k}\right)$ is then calculated as:

\begin{equation}
\mathcal{L}_{\mathrm{FM}}\left(G, D_{k}\right)=\mathbb{E}_{(\mathbf{s}, \mathbf{x})} \sum_{i=1}^{T} \frac{1}{N_{i}}\left[\left\|D_{k}^{(i)}(\mathbf{s}, \mathbf{x})-D_{k}^{(i)}(\mathbf{s}, G(\mathbf{s}))\right\|_{1}\right]
\end{equation}

where $T$ is the total number of layers and $N_{i}$ denotes the number of elements in each layer. The GAN discriminator feature matching loss is related to the perceptual loss, which has been shown to be useful for image super-resolution \cite{c28} and style transfer \cite{c29}. 

The full objective then combines both GAN loss and feature matching loss as a minimization of:

\begin{equation}
\left(\max _{D_{1}, D_{2}, D_{3}} \sum_{k=1,2,3} \mathcal{L}_{\mathrm{GAN}}\left(G, D_{k}\right)\right)+\lambda \sum_{k=1,2,3} \mathcal{L}_{\mathrm{FM}}\left(G, D_{k}\right)
\end{equation}

where the weight $\lambda$ controls the importance of the two terms. 

\section{Experimental Results}
\begin{figure*}
    \center
    \includegraphics[scale=0.47]{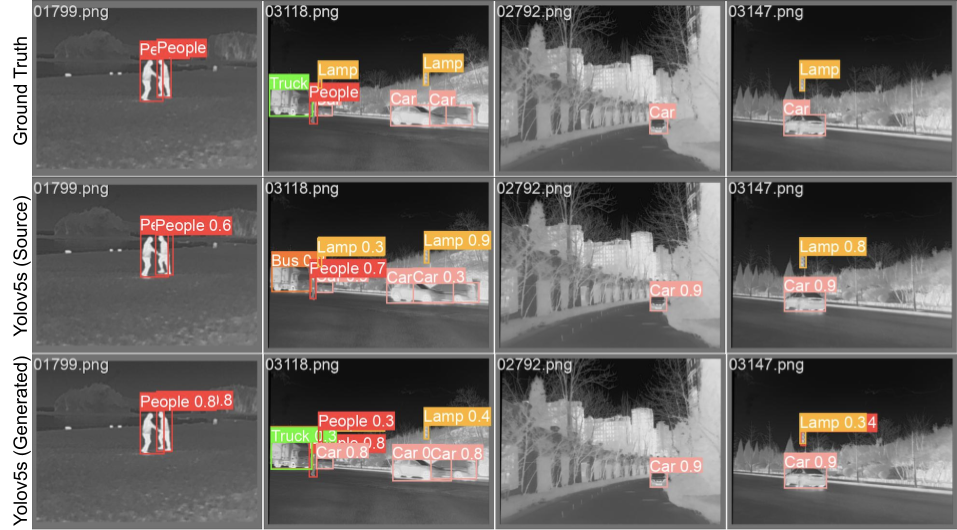}
    \caption{Visualization of predictions for a subset of images from the M3FD dataset. The first row represents ground truth bounding boxes, while the second row and third row represent Yolov5s model predictions on the test set when trained on the source training infrared images and generated images respectively. Clearly, training on generated images results in higher confidence scores and lower rates of misclassification.}
    \label{fig:Vis}
\end{figure*}
\begin{figure}
    \includegraphics[scale=0.42]{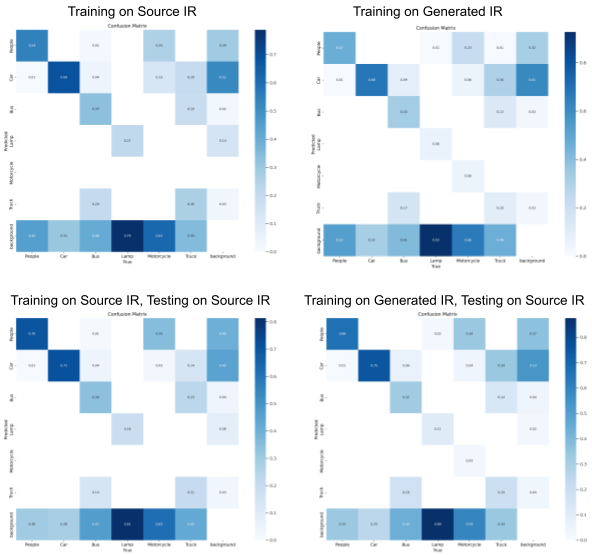}
    \caption[width=\columnwidth]{The confusion matrices for training and testing (only on source test set) for the M3FD dataset are similar, but notably, model performance is lower when training on generated IR but higher during testing. It is hypothesized that the generated infrared images, being marginally different from the source infrared images, force model generalization as compared to the potentially overfitted source-trained model.}
    \label{fig:Conf_Matrices}
\end{figure}
\subsection{Experimental Setup}
We evaluate the efficacy of generated images as training data against the source infrared baseline. All models are trained on the the same number of images as training distribution and evaluated on the official test split. For object detection, the evaluation metric used is the mean Average Precision (mAP).\\
\textbf{Dataset.} The M3FD EO-IR benchmark dataset \cite {c5} is used for all experiments involving both the generative model as well as the object detection models. It contains 4,200 co-registered EO-IR image pairs, the vast majority of which are of resolution 1024 x 768. There are 6 categories in total. 3,600 images are used for training, and 600 for testing.\\
\textbf{Implementation Details.} Yolov5s, Mask-RCNN and Faster-RCNN models are used to perform object detection. All models are pretrained on ImageNet, and the two RCNN models make use of ResNet-101 backbone. Yolov5s is trained for 100 epochs with a batch size of 16. With a probability of 0.5, augmentation techniques including scaling, transformation and rotation are used. Both Mask and Faster-RCNN are trained for 24 epochs with a batch size of 4, using a polynomial scheduler and a learning rate of $10^{-3}$, and use no augmentations. The SGD optimizer is used as a standard. Experiments were performed on an NVIDIA Quadro RTX 5000 GPU.

\subsection{Quantitative Analysis}
There is no universally accepted evaluation metric to evaluate the effectiveness of image generation methods. However, as the aim of our image translation network is to generate images suitable to train models on downstream tasks including object detection, we evaluate Yolov5s \cite{c3}, Mask-RCNN \cite{c30} and Faster-RCNN \cite{c31} on the original M3FD infrared testing set. Results are provided in TABLE \ref{table:M3FD}. All models demonstrate improved performance when trained on the dataset generated by the generative model, particularly when augmented with super-resolution \cite{c28}. Yolov5s improves by 4.6\%, Mask-RCNN by 5.27\% and Faster-RCNN by 4.291\%. We also provide quantitative metrics for analysis, Structural Similarity Index (SSIM) and Peak Signal-to-Noise Ratio (PSNR) in TABLE \ref{table:Metrics} and qualitative analysis against the baseline.

The model is also compared to the baseline and other methods on the basis of commonly used quantitative metrics for image processing. The metrics used are include peak signal-to-noise ratio (PSNR) and  structural similarity index (SSIM). For both metrics, higher values are indicative of a better quality of generated images. Average scores are presented in TABLE \ref{table:Metrics}.

\begin{table}
\caption{Performance comparison of object detection models of mAP (\%) using training data generated from the M3FD dataset.}
\centering
\begin{tabular}{|c|c|c|c|}
\hline
\textbf{I2I Translation Model}        & \textbf{YOLOv5s} & \textbf{Mask-RCNN} & \textbf{Faster-RCNN}       \\ \hline
Baseline          & 20.8   & \underline{27.128} & \underline{35.192}           \\ \hline
Proposed    & \textbf{26.1}    & 24.831 & 33.732 \\ \hline
Proposed + Super-Res          & \underline{25.4} & \textbf{32.398} & \textbf{39.483} \\ \hline
\end{tabular}
\label{table:M3FD}
\end{table}

\begin{table}
\caption{Quantitative Metrics for Proposed Methods: SSIM and PSNR.}
\centering
\begin{tabular}{|c|c|c|}
\hline
\textbf{I2I Translation Model}        & \textbf{SSIM} & \textbf{PSNR}      \\ \hline
Proposed & 0.853 & 27.950 \\ \hline
Proposed + Super-Res & 0.826 & 26.690 \\ \hline
\end{tabular}
\label{table:Metrics}
\end{table}

\subsection{Qualitative Analysis}
The predictions obtained from inferencing the Yolov5s model when trained on source infrared images and generated infrared images are compared against ground truths in Fig. \ref{fig:Vis}. All images are obtained from the test split of the M3FD dataset. The model trained on generated images demonstrates higher confidence scores on persons and cars, two of the most commonly occurring classes, and fewer misclassifications on less-represented classes, such as trucks and buses.

\section{Ablation Study}

\begin{table}
\caption{Validation for inverse mapping: performances across models trained on generated visible images and tested on infrared.}
\centering
\begin{tabular}{|c|c|c|}
\hline
\textbf{I2I Translation Model}        & \textbf{YOLOv5s} & \textbf{Mask-RCNN}       \\ \hline
Baseline          & 8.3   & 6.056          \\ \hline
Proposed    & \textbf{20.0}    & \textbf{7.062} \\ \hline
\end{tabular}
\label{table:VAL}
\end{table}

To validate the hypothesis that generated images ameliorate model generalization to unseen domains, we use the same GAN network to learn the inverse mapping from infrared to visible images. Results as seen in TABLE \ref{table:VAL} confirm the validity of our hypothesis.

\section{Conclusion}

In this paper, a novel two-stage technique is proposed to augment the accuracy of downstream computer vision tasks including object detection. In particular, the problem is formulated akin to the task of image translation, where a GAN is used to learn a bijective mapping from the visible to the infrared domain. Images so generated are more useful to object detection models, including Yolov5s, Mask and Faster R-CNN. We also investigate the usefulness of super-resolution in such an image augmentation pipeline. In conclusion, our method produces realistic and high-resolution images, retaining characteristics from visible images while learning the texture of infrared images. However, as a two-stage network, our model is unsuitable to real-time applications. Potential for further investigation through an end-to-end translation and detection network and other related strategies is left for future work.

\end{document}